\documentclass[10pt,twocolumn,letterpaper]{article}

\usepackage[pagenumbers]{cvpr} 

\usepackage{graphicx}
\usepackage{amsmath}
\usepackage{amssymb}
\usepackage{booktabs}
\usepackage{pythonhighlight}
\usepackage{cuted}

\usepackage{xcolor}

\usepackage[pagebackref,breaklinks,colorlinks]{hyperref}

\usepackage[capitalize]{cleveref}
\crefname{section}{Sec.}{Secs.}
\Crefname{section}{Section}{Sections}
\Crefname{table}{Table}{Tables}
\crefname{table}{Tab.}{Tabs.}

\begin{document}

\title{MapReader: A Computer Vision Pipeline for the Semantic Exploration of Maps at Scale}

\author{
Kasra Hosseini \quad 
Daniel C.S. Wilson  \quad
Kaspar Beelen \quad 
Katherine McDonough 
\\
The Alan Turing Institute, UK\\
{\tt\small \{khosseini, dwilson, kbeelen, kmcdonough\}@turing.ac.uk}
}

\maketitle

\begin{abstract}
 We present \textit{MapReader}, a free, open-source software library written in Python for analyzing large map collections (scanned or born-digital). This library transforms the way historians can use maps by turning extensive, homogeneous map sets into searchable primary sources. \textit{MapReader} allows users with little or no computer vision expertise to i) retrieve maps via web-servers; ii) preprocess and divide them into patches; iii) annotate patches; iv) train, fine-tune, and evaluate deep neural network models; and v) create structured data about map content. We demonstrate how \textit{MapReader} enables historians to interpret a collection of $\approx$16K nineteenth-century Ordnance Survey map sheets ($\approx$30.5M patches), foregrounding the challenge of translating visual markers into machine-readable data. We present a case study focusing on British rail infrastructure and buildings as depicted on these maps. We also show how the outputs from the \textit{MapReader} pipeline can be linked to other, external datasets, which we use to evaluate as well as enrich and interpret the results. We release $\approx$62K manually annotated patches used here for training and evaluating the models.
\end{abstract}

\section{Introduction}

\begin{figure*}[t]
    \centering
    \includegraphics[width=1.0\textwidth]{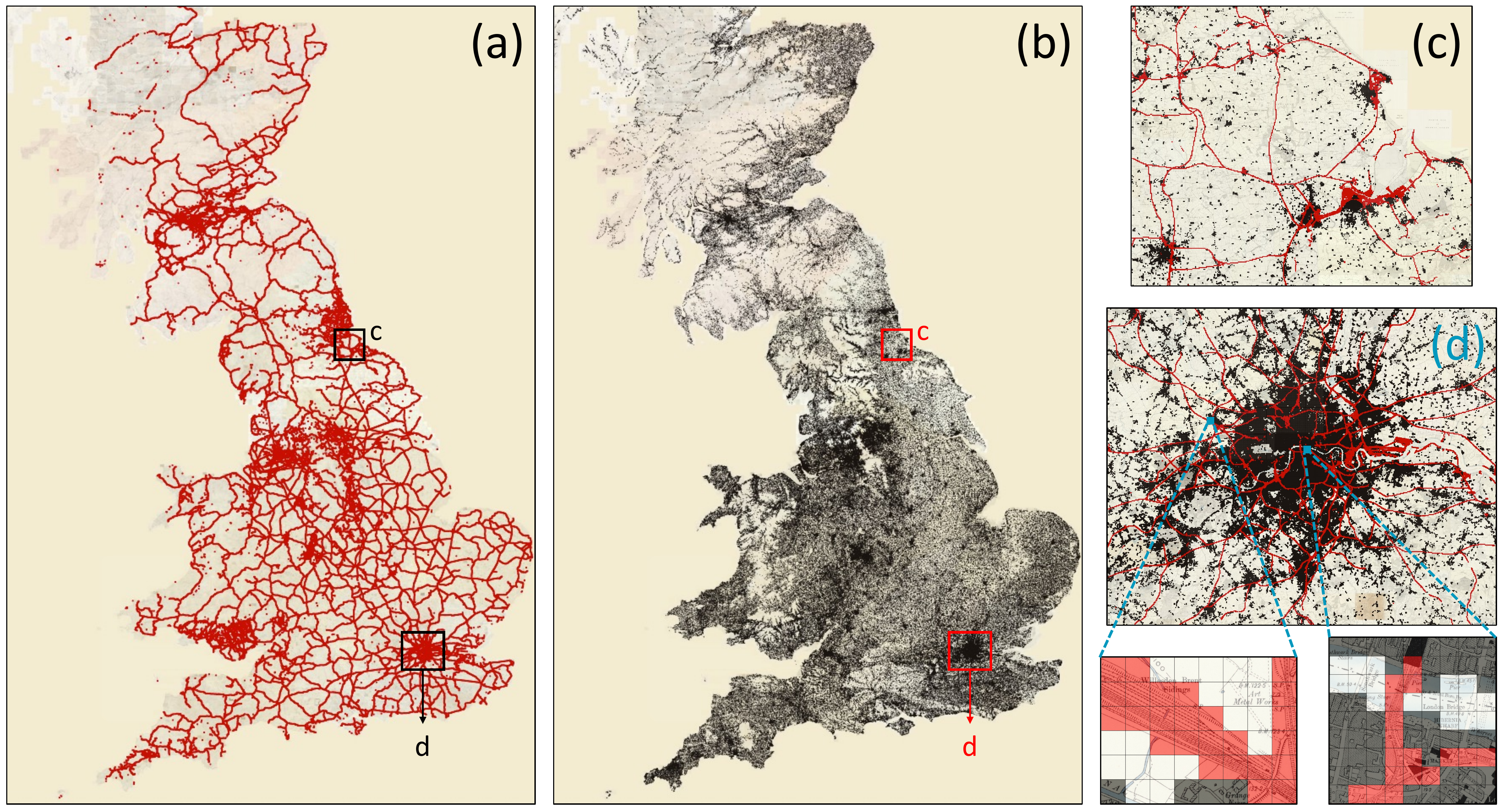}
    \caption{
  British railspace and buildings as predicted by a \textit{MapReader} computer vision model. 
  $\approx$30.5M patches from $\approx$16K nineteenth-century OS map sheets were used.
  (a) Predicted railspace;
  (b) predicted buildings;
  (c) and (d) predicted railspace (red) and buildings (black) in and around Middlesbrough and London, respectively.
  \textit{MapReader} extracts information from large images or a set of images at a patch-level, as depicted in the figure insets.
  For both railspace and buildings, 
  we removed those patches that had no other neighboring patches with the same label in the distance of 250 meters.
    }
\label{fig:mr_example_output}
\end{figure*}

Heritage collections of large map series contain immensely detailed historical and geographical information, but the contents of these maps have until recently remained out of reach for scholars wishing to perform spatial analysis. Initiatives for scanning and georeferencing map collections have brought such material into the digital realm – allowing researchers to browse them easily on screen, sometimes ``stitched'' together as a continuous layer. However, it remains impossible to search within these maps using the kind of systematic digital approach needed by historians or geographers to address questions in their fields. 

Our focus in this paper is on Britain's Ordnance Survey (OS), which in the nineteenth century produced definitive maps of the nation at three different scales. A single series of maps printed at a scale of 6'' to 1 mile, for example, contains $\approx$16K sheets. Newly available in digital form, these maps consist of a dataset which is both very large and very rich, with the potential for addressing many open questions in historical research. One example of such a research area relates to the arrival of railways in the British landscape, an issue we explore using the method outlined in this paper. To investigate qualitative geographical phenomena computationally requires researchers to translate, in this case, a visual map feature into a machine-readable label. This procedure is at the heart of \textit{MapReader}, which introduces what we term a ``patchwork method'' which divides our source image into constituent squares or ``patches'' which are then annotated according to the presence of a target feature. This differs from vector-based or segmentation approaches to visual source material and has certain affordances which are well suited to forms of spatial analysis that involve relational, additive or abstract target concepts, such as those typically of interest to humanists. We apply \textit{MapReader} to OS maps, but it can be used to search within any series, or very large, homogeneous set of, maps.

Any annotation process is a form of translation which inevitably involves a loss of descriptive information, which some have called a ``semantic gap''~\cite{arnold2019distant}. The present method helps address this problem by allowing flexible experimentation with decisions around the labelling and annotation of data. This method, in combination with these sources in particular, allows a wholly new approach to the fields of history and historical geography, framing questions it has not previously been possible to ask, for example, by thinking about visual typologies and suggesting abstract spatial relationships that do not yet exist in the literature. 

In this study, we select annotation labels, or categories of historical information, and identify them at the ``patch'', rather than pixel, level (see Fig.~\ref{fig:mr_example_output}). Patches are meaningful semantic units that are highly flexible both in nature (their size can vary) and as part of a workflow (they can remain stable while labels change or are tested iteratively).
Patches can be annotated from one or more maps without a need to annotate one whole sheet. 
This paves the way to design efficient annotation mechanisms compared to, for example, image segmentation methods.
While well-known in computer vision (CV) research, particularly in  medical imaging~\cite{MCCOMBE20214840}, we present patches as a novel structure for historical spatial data found on maps.

We demonstrate how \textit{MapReader} (a) enables researchers to create and refine annotation schema and provides procedures for (b) digital annotation and (c) model training that ensure reliable out-of-sample generalization. The power of \textit{MapReader} lies in its ability to analyze at high volumes the products of major (series) mapping initiatives, which produced thousands of maps around the world between the late eighteenth and late twentieth centuries. Its interactive annotation tool can be used by researchers to label a large number of patches. \textit{MapReader} provides functionalities to train or fine-tune various CV models using the annotated datasets. The main contributions of this paper are as follows:

\begin{itemize}
    \item An open-source, end-to-end CV pipeline for the semantic exploration of high-volume map collections using the patch method (see Fig.~\ref{fig:pipeline}).\footnote{
    The \textit{MapReader} library is released under MIT License.
    Its source codes are on GitHub (\url{https://github.com/Living-with-machines/MapReader}). 
    We have also provided Jupyter Notebooks on GitHub to reproduce the main results presented in this paper.
    } 
    \item A new, expert-labelled training and evaluation dataset for the CV classification task, consisting of 62,020 human-annotated patches.
    \item First nation-wide analysis of historical maps of nineteenth-century Britain. This task consists of classifying $\approx$30.5M patches from $\approx$16K maps.
    \item A rigorous analysis of visual annotation and model training for the semantic exploration of maps at high volumes. We focus on an initial interpretation of the results for a specific case study: the relationship between rail and the built environment in late nineteenth-century Britain.
\end{itemize}

\section{Related work}

Historical map processing is a growing subfield in GIScience that has thus far combined the methods of image processing with the concerns of geographers. Using rule-based and now machine-learning methods, significant advances have been made in automating the otherwise tedious processes of generating GIS vector and raster data from images~\cite{budig_extracting_2018, chiang_historical_2020}. Researchers are working to improve methods for creating datasets of, for example, building footprints\cite{hecht_mapping_2019} and road and rail networks~\cite{chiang_strabo:_2010, chiang_training_2020} using very large, homogeneous series maps comparable to OS maps, such as the United States Geological Service topographical maps.

This research often builds on the lessons of working with remote sensing imagery or aerial photography to classify map content, relying on the spatial data structures of GIS to translate landscapes into machine-readable content. ``Mining'' maps in this way can produce useful data for studying, for example, transport costs or settlement patterns~\cite{uhl_urban_2020, uhl_fine-grained_2021, leyk_two_2020}. New work in the Digital Humanities exploring the semantic segmentation of maps \cite{petitpierre_generic_2021} contributes to general (e.g., cross-collection) solutions for automating the creation of vector data. Our method complements these ways of working with historical maps that are now available as scanned images, but it is also distinct in its approach to this source material.

While historians and others have written extensively about the spatial character of British industrialization, many questions remain unanswered \cite{floud_transport_2014}. This is in part because the heterogeneity of historical sources makes it difficult to compile evidence at a high enough resolution of detail to assess national trends in the built environment. OS maps have the potential to overcome this shortcoming and have been used by scholars to (manually) identify chimneys\cite{heblich_east-side_2020}  and (automatically) extract land use types\cite{baily_extraction_2007}. More often, however, maps have been used as figures to illustrate a specific point in a historian's argument, rather than as sources of new information in themselves. 

Our approach differs from previous map processing work because we integrate the insights of scholars working with historical maps in a non-digital setting, including ideas of source criticism and data provenance, in this case in relation to the contested nature of map production itself~\cite{wigen_time_2020, harley_silences_1988, edney_cartography:_2019, akerman_decolonizing_2017}. It is also our explicit ambition to bring historical maps into dialogue with other datasets, for example, texts and other sources of information about the past.
Scholars working in the Digital Humanities have only recently embraced a ``visual digital turn''~\cite{wevers2020visual}, which has signalled a welcome effort to reflect on the theoretical and practical implications of working with visual data at scale. Arnold and Tilton \cite{arnold2019distant} introduced ``distant viewing'' as a research methodology which critically interrogates the ``interpretive nature of extracting semantic metadata from images''. Our work builds on these observations, in particular, in the way that our interface enables scholars to fine-tune annotations and models iteratively, in constant dialogue with historical materials, in order that its users remain alert to the dangers of the ``semantic gap'' identified in this literature.

\section{M\lowercase{ap}R\lowercase{eader} pipeline}
\label{sec:mapreader_pipeline}

\begin{figure}[t]
    \centering
    \includegraphics[width=1.0\linewidth]{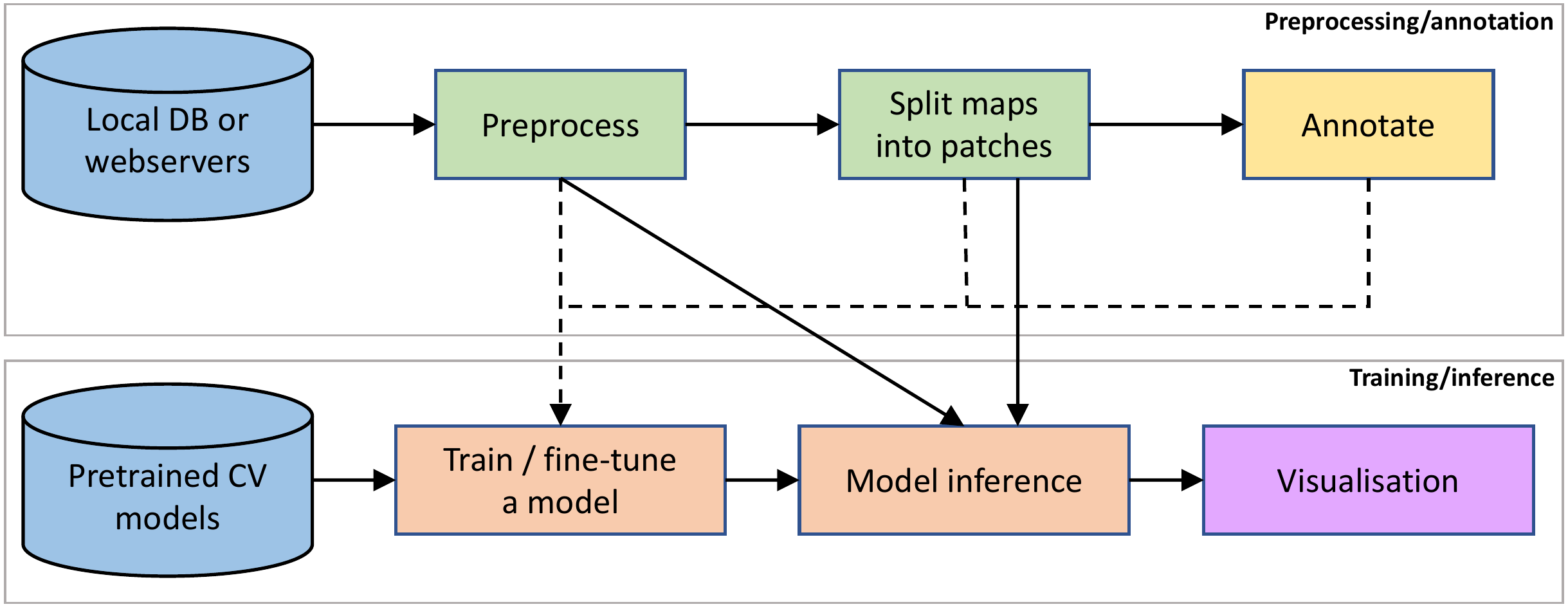}
    \caption{\textit{MapReader} is an end-to-end CV pipeline with two main components: preprocessing/annotation (top) and training/inference (bottom).}
\label{fig:pipeline}
\end{figure}

Fig.~\ref{fig:pipeline} shows the main components of the \textit{MapReader} pipeline. The first component, ``preprocessing/annotation'', has functionalities to load locally stored images or to retrieve maps via webservers. 
Here, we use the National Library of Scotland (NLS) Historical Maps API to download $\approx$16K nineteenth-century OS map sheets and their metadata. 
The same interface can be used to download maps from other tileservers (e.g., tiles based on the OpenStreetMap data). 
\textit{MapReader} can then be used to preprocess the retrieved maps, such as resampling the images, removing borders outside the neatline or reprojecting the map.

In the next step, preprocessed map sheets are sliced into patches. The size of these patches can be specified by the number of pixels or by length in meters. For the latter, the geographic information (i.e., latitudes and longitudes) of the bounding box corners must be available in the map file or metadata. Determining the patch size is an important step in the \textit{MapReader} pipeline: it should be large enough that the target label is presented in one patch with enough visual features to be identifiable by both the human annotator and the CV classifier; but it should be small enough to offer good spatial resolution.

\textit{MapReader} provides an interactive annotation tool to label patches. Its interface shows a patch to be annotated and, if selected by the user, a visual context around that patch (refer to Appendix~\ref{app:annot_tool} and  Fig.~\ref{fig:mr_annotation_interface} for an example). The user has control of the labels, the size of the context image and the selection criteria (by default, the patches are selected randomly, but they can be ordered by the mean or standard deviations of pixel intensities). 
We developed this interface based on the feedback from researchers with different backgrounds, such as historians, linguists, data scientists and software engineers. We have tested the usability of the annotation tool and its user experience in various tasks. \textit{MapReader}'s annotation tool also supports reviewing model predictions and adding the (human expert) revised labels to the gold standard. We used this approach to annotate 62,020 patches (see Section~\ref{subsec:annotations} for details).

The second component, ``training/inference'' (Fig.~\ref{fig:pipeline}, bottom), has functionalities to read in the annotations and to split them into three sets: train, validation and test. We use a stratified method for splitting the annotations, that is, each set contains approximately the same percentage of samples of each target label as the original set \cite{scikit-learn}. Next, the user defines transformations which will be used for data augmentation during training. 

The ``training/inference'' component of \textit{MapReader} is built upon \texttt{torchvision} and \texttt{PyTorch} \cite{paszke2019pytorch} and has been tested on both CPU and GPU. A custom defined CV model or a pretrained model\footnote{Various pretrained models can be downloaded from \texttt{torchvision} or \texttt{PyTorch Image Models} \cite{rw2019timm}.} can be used in this module. 
\textit{MapReader} can also be used to build model ensembles. We will discuss this functionality, ``context-aware patchwork method'', in Section~\ref{subsec:patch_level_pros_cons}.
The user can define (or use existing) optimizers, schedulers and optimization criteria. 
\textit{MapReader} allows the user to ``freeze'' any layers in the model architecture when training or finetuning a model. One learning rate can be assigned to all layers, or \textit{MapReader} can assign different learning rates to each layer\footnote{The layer-wise learning rate was inspired by the implementation in fastai~\cite{Howard_2020}.}. 
During training and evaluation, \textit{MapReader} collects some performance metrics which can be plotted in real-time using TensorBoard \cite{tensorflow2015-whitepaper} or offline. These metrics include training and validation losses, F1-scores (for each label separately as well as micro and macro averages), accuracy, precision and recall. 
Moreover, \textit{MapReader} provides several plotting tools to visualize and analyze the results of model inference.

\textit{MapReader} is available as a Python library.
We provide extensive documentation, including examples for the main functionalities of \textit{MapReader} in Jupyter Notebooks, to enable the smooth adoption of its components.
All its functionalities, such as retrieving or loading maps, slicing them into patches, reading annotations, training, fine-tuning and model evaluation, have easy-to-use interfaces. Consult Appendix~\ref{app:mr_python_interface} and \textit{MapReader}'s GitHub page for additional information and examples.

\section{Experiments and results}

\subsection{Dataset} 
\label{subsec:dataset}

We used the 2nd edition of the OS 6'' to 1 mile maps (1:10,560 scale). 
These sheets were surveyed between the late 1880s (for Scotland) and early 1890s (for England and Wales) and the beginning of World War I (first at 6'', and after 1893, at 25'' to 1 mile) and printed (and re-printed, as revisions were made), at the reduced scale of 6'', between 1888 and 1913.\footnote{Distinguishing between the 1st and 2nd edition maps can be artificial, but the vast majority of sheets known as 2nd edition sheets were prepared at the beginning of the 1880s.}

Of all the sheets for England, Wales and Scotland scoped for the 6'' series, only about 0.1\% (or 15 sheets, all in England or Wales) from the 2nd edition have not been scanned by the NLS. The scanned sheets have been made available as georeferenced images with content outside the neatline masked. The NLS maintains a seamless layer for this edition, which we access via their tileserver.\footnote{Details about this layer are available at \url{https://maps.nls.uk/projects/subscription-api/\#gb6inch}.}

We downloaded 16,439 map sheets ($\approx$600G) from the NLS tileserver, and sheet-level metadata was shared by the NLS (and can be accessed via \textit{MapReader}'s repository). This metadata is essential to understanding the complex temporalities of this edition (as evidenced by the spread of dates between the earliest and the latest sheets surveyed and printed). In addition to these dates, metadata includes different identifying information for each sheet,
whether a sheet is held by the NLS, whether it has been digitized, and more. 
After downloading the maps, we sliced each sheet into square patches of size $\approx$100m$\times$100m.\footnote{\textit{MapReader} estimates the width and height of each pixel by using coordinates of the borders.} This step generated $\approx$30.5M patches out of 16,439 map sheets, and it took $\approx$32 hours on 6 cores.

\subsection{Annotations}
\label{subsec:annotations}

We report on experiments to capture two kinds of physical structures that are important visual signals on the maps for rapidly changing nineteenth-century communities: buildings and railway infrastructure. Any patch containing any building or any railway infrastructure is labeled as such. Selecting ``building'' and ``railspace'' as our two basic labels resulted in the following annotation possibilities: (1) no [building or railspace]; (2) railspace; (3) building; and (4) railspace and [non railspace] building.
These two basic measures of development are an early demonstration of \textit{MapReader}'s utility for identifying, collating, analyzing (quantitatively) and viewing (qualitatively) the content on thousands of maps. Combining these outputs, we piece together a ``visual census'' of Britain~\cite{hosseini_maps_2021}.

\paragraph{Railspace}
For one of our labels, we introduce the concept of \textit{railspace} as a key element of the industrializing landscape. We are interested in understanding the impact of rail not simply as a means of transport. Railways include inter-related structures which not only impacted opportunities for industrialization and worker mobility, they also reshaped communities. 
\textit{Railspace} therefore contains anything represented on the map related to the use of rail - including single and double tracks, stations, depots, and embankments (but excluding urban trams). This concept goes beyond current railway track or station datasets to offer a comprehensive view of the footprint of rail across Britain at its height during the late nineteenth century.

\paragraph{Buildings}
Like \textit{railspace}, the building label captures another basic piece of information about the built environment of the nineteenth century. The rapid increase of building stock - industrial, residential, civic, and commercial - is an important indicator of the impact of industrialization on an area. With no access to open data about the location of buildings in the past across all of Britain, we developed this label as a coarse measure of the extent of buildings on the map.

\paragraph{Gold standard} 
The gold standard consists of 62,020 manually annotated patches. In our experiments, we used 37,212 patches for training, 12,404 for validation and 12,404 for testing. The dataset is imbalanced with the following distribution: 56,372, 1,041, 3,634 and 973 patches for the four labels stated above, respectively. We used a stratified method for splitting the annotations into training, validation and test sets, that is, each set contains approximately the same percentage of samples of each target label as the original set.

The patch-level approach allows us to collect large labelled datasets from one or more map sheets.
We will discuss this in more detail in Section~\ref{subsec:patch_level_pros_cons} (refer to ``Speeding up annotation'').

\subsection{Deep neural network classifiers}
\label{subsec:dl_net_class}

\begin{table*}
\caption{
Performance of CV classifiers on the test set with 12,404 labelled patches (20\% of all manually annotated patches). 
We used pretrained models from \textit{PyTorch Image Models} \cite{rw2019timm} and \textit{PyTorch} \cite{paszke2019pytorch}.
The models are ordered according to F1-macro. 
The F1-score for each label is also listed. 
F1-0: no [building or railspace]; F1-1: railspace; F1-2: building; and F1-3: railspace and [non railspace] building.
Time is for model inference on all 12,404 patches.
Refer to Appendix~\ref{app:compare_valid_set} and Table~\ref{table:models_valid_set} for the results on the validation set.
} \vspace{-0.5cm}
\begin{center}
\begin{tabular}{lrrrrrrlr}
\toprule
                     Model &  F1-macro &  F1-micro &  F1-0 &  F1-1 &  F1-2 &  F1-3 & Time (m) &  \#params (M) \\
\midrule
     resnest50d\_4s2x40d \cite{zhang2020resnest} &     \textbf{97.35} &     \textbf{99.55} & 99.81 & 96.40 & \textbf{97.41} & \textbf{95.79} &                         11 &         28.4 \\
            resnest101e &     97.17 &     \textbf{99.55} & \textbf{99.83} & \textbf{97.36} & 97.20 & 94.27 &                          8 &         46.2 \\
  swsl\_resnext101\_32x8d \cite{DBLP:journals/corr/abs-1905-00546} &     96.28 &     99.38 & 99.76 & 96.19 & 96.13 & 93.05 &                         19 &         86.8 \\
              resnet152 \cite{he2015deep} &     96.19 &     99.25 & 99.67 & 95.71 & 95.17 & 94.21 &                         10 &         58.2 \\
resnest101e\_no\_pretrain &     91.12 &     98.12 & 99.13 & 87.87 & 88.65 & 88.83 &                         12 &         46.2 \\
  tf\_efficientnet\_b3\_ns \cite{tan2020efficientnet} &     90.00 &     98.33 & 99.29 & 88.29 & 89.23 & 83.19 &                         11 &         10.7 \\
       swin\_base\_patch4 \cite{liu2021swin} &     76.92 &     93.32 & 96.54 & 64.52 & 67.19 & 79.43 &                         16 &         86.7 \\
   vit\_base\_patch16\_224 \cite{dosovitskiy2020image} &     62.50 &     91.03 & 95.77 & 37.93 & 62.93 & 53.37 &                         15 &         85.8 \\
\bottomrule
\end{tabular}
\end{center}
\label{table:test_models_01}
\end{table*}

We fine-tuned various CV models on the training set of our gold standard with 37,212 labelled patches (see Section~\ref{subsec:annotations}).
Our multiclass classification task has four classes as follows: no [building or railspace]; railspace; building; and railspace and [non railspace] building.

We followed the same training procedure for all the model architectures listed in Table~\ref{table:test_models_01}. 
Each patch has a size of $\approx$100m$\times$100m. (number of pixels varies depending on the size and zoom-level of a map sheet).
Before feeding the patches to a model, 
several steps of data transformations were applied (i.e., data augmentation), including horizontal and vertical flips, Gaussian blur and resizing to lower resolution images with 50$~\times$~50 pixels.
All of these transformations were applied randomly with a probability of 0.25.
The patches were then normalized using the mean and standard deviations of pixel intensities in all the patches in our dataset, and they were resized based on the CV model and its required input size (e.g., 224$\times$224 pixels in resnet or resnest models).

As discussed in Section~\ref{subsec:annotations}, we used a stratified method for splitting the annotations into training, validation and test sets. However due to the class imbalance in our gold standard, we employed a weighted sampler when generating training and validation batches. This weighted sampler increases the number of minority classes and tries to form a balanced dataset. 

We used Cross Entropy criterion and the AdamW~\cite{loshchilov2018decoupled} optimization method with a linear, layer-wise learning rate from 0.0001 for the first layer to 0.001 for the last layer in a neural network architecture.
In AdamW, we set $\beta_{1}$ and $\beta_{2}$, the coefficients used for computing running averages of gradient and its square, to 0.9 and 0.999, respectively.
We also used a scheduler to decay the learning rate of each parameter group by 0.1 every 5 epochs. 
We set the batch size to 32 and fine-tuned for 30 epochs over the training set.

For each neural network architecture, the model with the least validation loss was selected. 
We then compared the F1-macro scores of these selected models on the validation set with 12,404 labelled patches. 
(Refer to Appendix~\ref{app:compare_valid_set} and Table~\ref{table:models_valid_set} for more detail).
``Resnest101e'' had the highest F1-macro score of 96.74\% on the validation set (and we used this model for all the results presented in this paper).
Finally, we measured the performance of the selected models on the test set.
This step was done only once, and the results are reported in Table~\ref{table:test_models_01}.

\textit{MapReader} provides an easy-to-use interface to experiment with hyperparameters of different stages in the pipeline. Some of our experiments were not used in the final training as they did not improve the performance significantly. These included changing the weights of the weighted sampler from $10/(class\_count)$ to $1/(class\_count)$; switching off the scheduler; assigning learning rates using a geometric progression (i.e., spaced evenly on a log scale) from $0.00001$ for the first layer to $0.001$ for the last layer in the neural network architecture; and ``freezing'' all the layers except for the last during fine-tuning.

All the models in Table~\ref{table:test_models_01} were pretrained and further fine-tuned on our training set except ``resnest101e\_no\_pretrain''.
Comparing the two versions of ``resnest101e'' shows that the predictive performance was substantially improved by fine-tuning the pretrained model rather than training the model from scratch \cite{hinton2006reducing, larochelle2009exploring}.

Our selected model, ``resnest101e'', was then used for model inference on all 30,490,411 patches out of $\approx$16K map sheets covering England, Wales and Scotland. 
In practice, this step is trivial to process in parallel. 
We distributed the patches on four NVIDIA Tesla K80 GPUs, and it took $\approx$172 GPU hours in total.
This time includes model inference and some pixel-level calculations, such as mean and standard deviation pixel intensities at patch level. 
For model inference, we applied both normalization and resizing transformations as described above.

After performing model inference on all patches, one of the issues in our predictions was disconnected areas of railspace. Given our prior knowledge about the structure of rail networks, it would be safe to conclude that the disconnected areas are false positives. To take advantage of the structure present in nearby railspace/non-railspace labels, we introduced a post-processing step. The goal was to improve the prediction for a given map patch using nearby predictions. 

To do this, we first computed the distance of a patch to its closest neighboring patch with the same predicted label.
We used k-d tree to efficiently find neighbors and calculate distances \cite{k-d_tree}.
For both rail infrastructure and buildings, 
we then removed those patches which had no other neighboring patches with the same label in the distance of 250 meters.
This step removed 4,082 ``isolated'' patches ($\approx$0.8\%) out of ``487,360'' total patches for rail infrastructure.
We will share both the raw and postprocessed outputs.

\section{Discussion}

This section addresses challenges in designing a CV pipeline for humanistic, computational map analysis at scale, and the research opportunities that flow from the interface and experiments reported in this study.
We also discuss the limitations of our current method and future avenues to address them.

\subsection{Systematic errors}
\label{subsec:systematic_errors}

\begin{figure}
    \centering
    \includegraphics[width=1.0\linewidth]{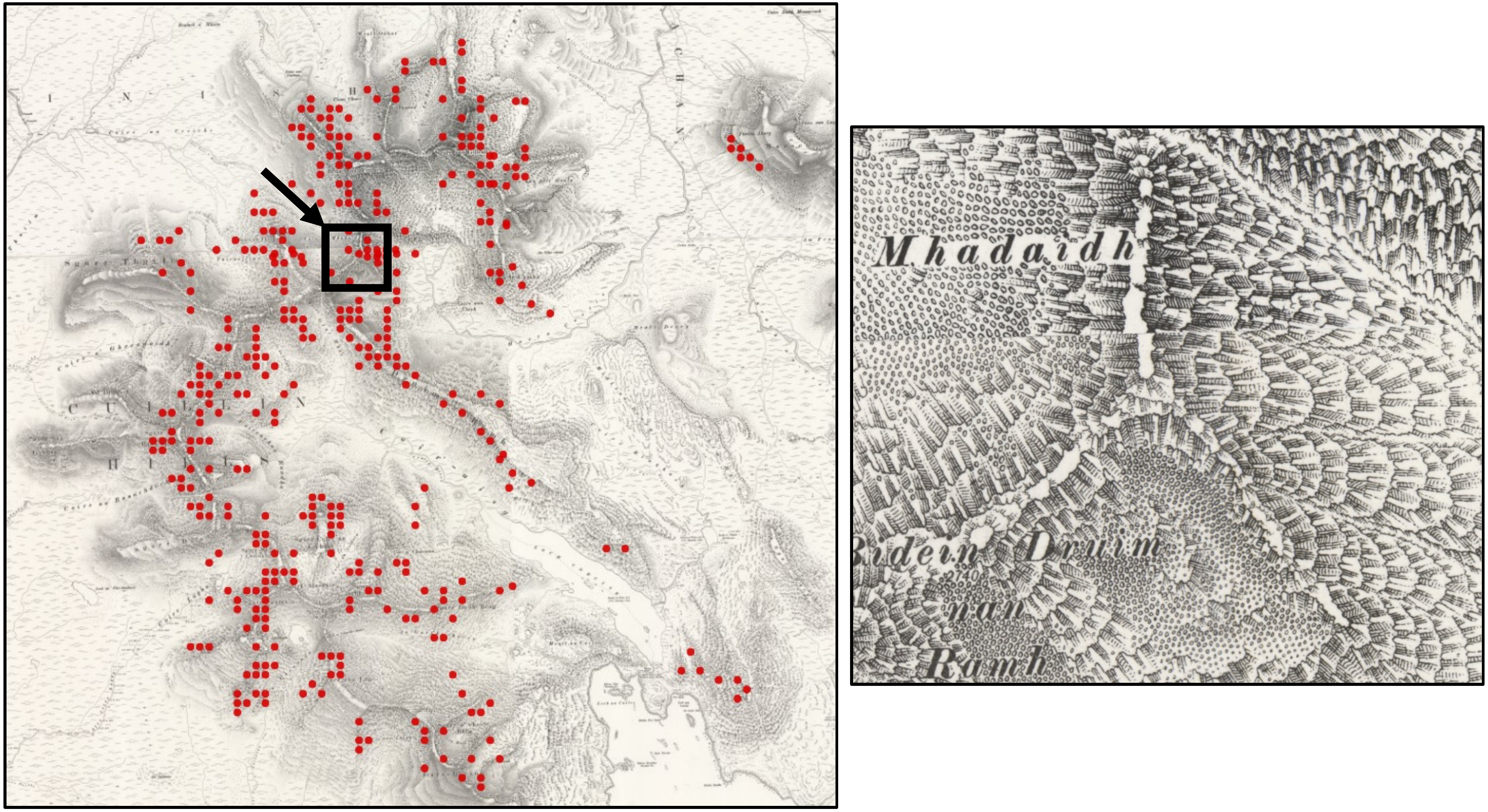}
    \caption{
    Systematic errors in predicting railspace (red dots). 
    The Cuillin Hills (Isle of Skye, Scotland) contain an example of features drawn on the map using hatched rectangles.
    This is similar to how rail tracks are represented on the map.
    The initial, ``weak'' model predicted these as railspace (false positive).
    See text for details.}
\label{fig:systematic_error}
\end{figure}

A major issue we observed in our experiments was that high test scores achieved on gold standard data mask poor performance outside of the annotated sample due to unobserved biases in the data. 
We explain how overcoming this issue requires a combination of annotation methods \textit{and} training procedures. 
An affordance of working with maps as data is that errors are straightforward to spot (by visually inspecting the results stitched together on a map at a regional, or even national, level), which allows for further refinement of both annotation and models.

In one of our initial experiments, 
we fine-tuned a ``resnest101e'' model on 7,766 patches.
All the preprocessing and training steps were the same as our final model (described in Section~\ref{subsec:dl_net_class})
except the number of patches used for training (7,766 patches versus 37,212 in the final model).

This ``weak'' model had a F1-macro of 95.96\% on the validation set with 2,589 patches.\footnote{Other performance metrics of this model are F1-micro = 99.19\%; F1-0 = 99.64\%; F1-1 = 92.31\%; F1-2 = 98.16\%; F1-3 = 93.75\%.}
After performing the model inference on all $\approx$30.5M patches, we visually inspected the outputs on maps similar to what is shown in Fig.~\ref{fig:mr_example_output}.\footnote{We used \textit{kepler.gl} (\url{https://github.com/keplergl/kepler.gl}) for visualization and qualitative model inspection.}
This revealed some systematic errors in the outputs of our model. 
Fig.~\ref{fig:systematic_error} shows one example for this in Scotland where rocks were predicted by the  ``weak'' model as buildings and sometimes railspace because of the use of hatched lines that mimic both terraced housing and railway tracks. Hachures like this were used prolifically by OS draftsmen in the Highlands and along the cliffs of Britain's undeveloped coasts.
Other common errors after the first round of training included coastal walls, reservoirs, drainage ditches, river or lake shores, and some rocky areas predicted as railspace patches.

False negatives (e.g., missed railspace patches) included railways passing through forests and railways on the full-size Scottish sheets. 
In general, the full-size sheets in Scotland threw up more systematic errors than the more common quarter-size sheets (in England and Wales). This motivated switching our definition of patch size from 100$\times$100 pixels to $\approx$100m$\times$100m: each patch now represented the same size area and this did indeed improve prediction of railspace and buildings across the areas of Scotland printed on full sheets. Further systematic errors include false negatives for isolated farms and other buildings. 
We increased the dataset to include more sheets with a high percentage of rural space to improve accuracy in this area.
There are well-founded concerns among humanists\cite{da_2019} (though few from historians) for making arguments based on inferred data, and we know that we cannot account for all errors. However, given the high quality of our results, we find this limitation to be minimal. Furthermore, we will continue to assess the effect of errors on interpretation as we conduct new experiments.

\subsection{Linking external data}

For each patch, \textit{MapReader} outputs a predicted label and its confidence score (see Section~\ref{subsec:dl_net_class}). 
Each map sheet also contains some metadata, including geographic coordinates of its corners, and survey and publication dates.
Using geographic coordinates of a map sheet, therefore, we can compute the center latitude and longitude of each patch,
which can be used to plot the results on a map (see, for example, Fig.~\ref{fig:mr_example_output}).
They can also be used to link the outputs to other, external datasets.
We use this to evaluate as well as enrich the results of \textit{MapReader}.

One dataset directly related to our work is \textit{StopsGB}~\cite{lwm-station-to-station-2021}, a structured and linked dataset of over 12,000 railway stations.
We used this dataset to evaluate the results of \textit{MapReader}.
The stations listed in \textit{StopsGB} were filtered by the survey dates of our map sheets such that railway stations would be expected to appear on a given map sheet, and we then computed the distance from each station to the center of its closest railspace patch. 89.3\% of \textit{StopsGB} stations were within 150 meters of the center of a railspace patch (each patch is $\approx$100m$\times$100m). 
The discrepancy between the two datasets could either be due to errors in \textit{MapReader}'s predictions or errors in \textit{StopsGB}. 
(Refer to Appendix~\ref{app:linking_stops} and Fig.~\ref{fig:mr_cv_quicks} for a visual comparison between the two datasets.)
Our qualitative comparison shows that most of the stations appearing far from any railspace patch are caused by errors in the \textit{StopsGB} dataset.

\subsection{Digital History Research}

In this section, we briefly discuss the potential of \textit{MapReader} as a research tool for (digital) History, focusing on the distribution of railspace in the built environment as a case in point.
The predicted patches produced by \textit{MapReader} help historians locate objects in the landscape, but, as we demonstrate here, the labels themselves also enable scholars to construct novel concepts for studying place by virtue of the way they can be combined either with each other or with other geospatial datasets.
For example, we used ``buildings'' and ``railspace'' patches in combination to explore the way the two features intertwine on the maps and, specifically, focusing on areas where rail appears to have ``dominated'' its surroundings.
By taking into account the spatial context of each patch, we can enhance the spatial semantics of our analysis and identify space in terms of the \textit{relation} between (and not just the presence of) our categories of interest.
Fig.~\ref{fig:london_railspace} centers on London showing building patches with more than 20\% of railspace patches within a radius of 500 meters. 
While this figure partly confirms what is known about railspace in London, \textit{MapReader} allows us to go beyond well-known cases like these to work at the national scale: we can quantify every patch by its rail density, thus applying the same metric to the whole country. Such a perspective enables historians to search for places and compare them along new axes; for instance, by identifying all areas that resemble the patterns of London railspace (as in Fig.~\ref{fig:england_railspace}). While London is unique in its size, in terms of its railspace clusters it has similarities with industrialized northwestern and northern cities, which are highlighted at the top of the map in Fig.~\ref{fig:england_railspace}. More specifically, Fig.~\ref{fig:leeds_railspace} shows which areas in Leeds have the same railspace-dominant properties as the London Docklands.

\subsection{Patchwork method: advantages and limitations}
\label{subsec:patch_level_pros_cons}

\paragraph{Speeding up annotation.} The patchwork method allows us to collect labelled data, from one or more maps, without needing to annotate every patch on one map sheet. 
This provides a convenient method to extend the labelled dataset. 
We used this at two points in our pipeline. 
First, we trained a ``weak'' classifier with $\approx$10K manually annotated patches and used it for model inference on all map sheets (with $\approx$~30.5M patches). 
As discussed in Section~\ref{subsec:systematic_errors}, this helped us to identify (systematic) errors and biases in the data, and subsequently, to better extend the labelled dataset by sampling from regions where the ``weak'' model made most of the mistakes (by visually inspecting the results and identifying false positives and false negatives).
Second, instead of annotating the patches from ``scratch'', we asked the annotators to review the predictions of the initial, ``weak'' classifier. 
Large areas on maps can be blank or have visual features that can be correctly identified even by a ``weak'' classifier.
Instead of spending time annotating such patches, the annotator can quickly review them in batches and spend time on more challenging cases.
The revised labels are then added to the gold standard.

\begin{figure}
    \centering
    \includegraphics[width=1.0\linewidth]{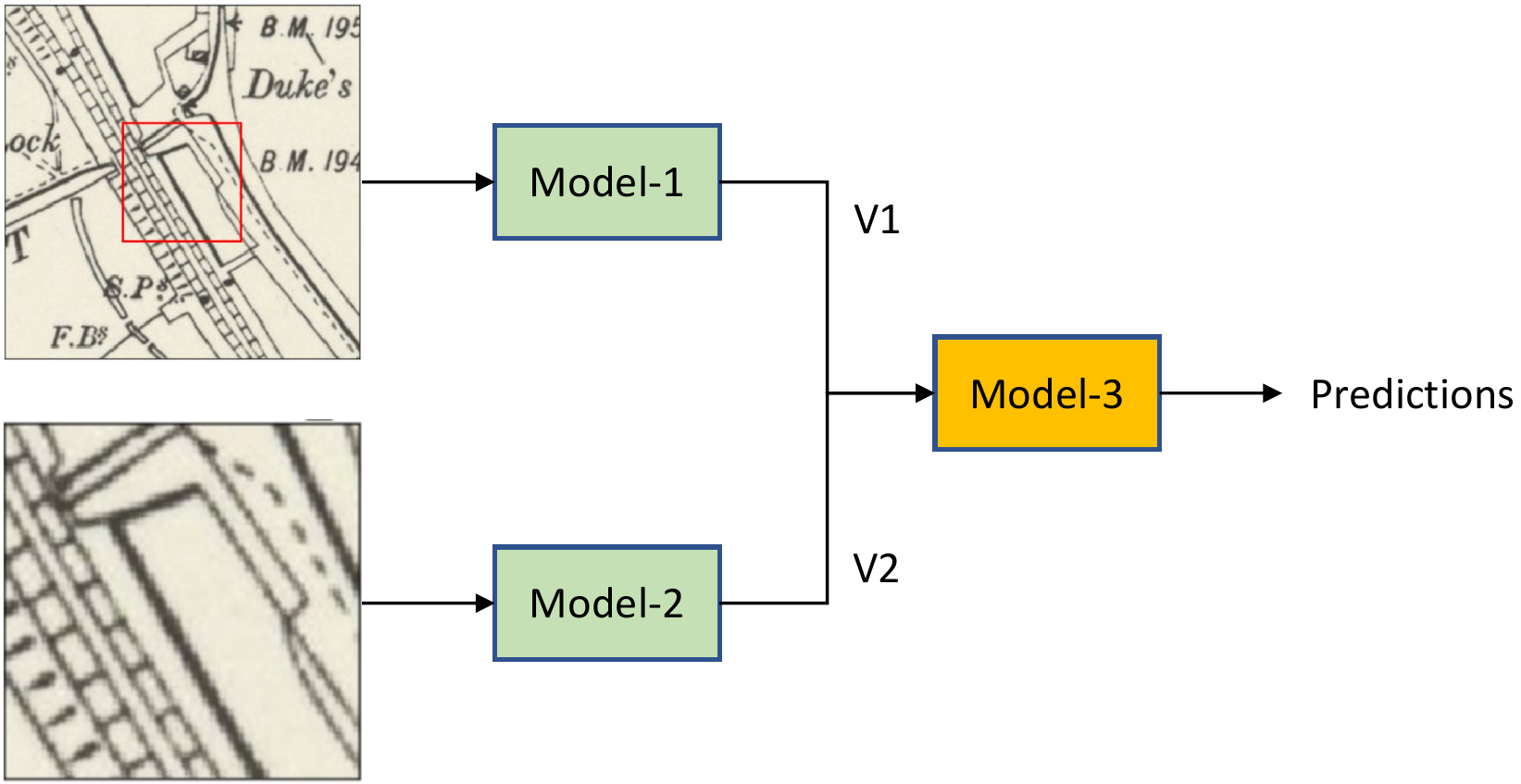}
    \caption{
    Model ensembles in \textit{MapReader}. 
    A patch (bottom) and its context image (top) are fed into two CV models, Model-1 and Model-2.
    A combination of the vector representations generated by these two models ($V1$ and $V2$) is given as input to Model-3 to calculate the final output (or prediction).
    See text for details.
    Red square in the context image marks the bounding box of the target patch and is added for visual reference in this figure.
    }
\label{fig:ensemble_models}
\end{figure}

\paragraph{Context-aware patchwork method.} 
We considered two factors to select the size of patches in our experiments as described in Section~\ref{sec:mapreader_pipeline}.
However, one of the limitations of the patchwork method is that the neighboring patches are not used when training or performing model inference.

To capture the strong dependencies between the neighboring patches, \textit{MapReader} supports building model ensembles as shown in Fig.~\ref{fig:ensemble_models}. 
\textit{Model-1} and \textit{Model-2} are two CV models with neural network architectures defined by the user.
These models can have different or similar architectures. For example,
in one of our experiments, \textit{Model-2} was a pre-trained model on our patch dataset while 
\textit{Model-1} was a pre-trained model from \textit{PyTorch Image Models} \cite{rw2019timm} with a different architecture.
As shown in Fig.~\ref{fig:ensemble_models},
a patch and its context image are fed into \textit{Model-2} and \textit{Model-1}, respectively.
In practice, the user only specifies the size of the context image, and \textit{MapReader} extracts and preprocesses the context image from the dataset.
\textit{Model-1} and \textit{Model-2} generate vector representations for the input images, \textit{V1} and \textit{V2}.
The size of these vectors are defined by the user.
A combination of these two vectors (e.g., by concatenation) is then fed into \textit{Model-3} for prediction.

Such model ensembles can be an efficient approach to achieve high-performing CV models \cite{wang2021wisdom}. 
As described above, \textit{MapReader} already has the functionality to build model ensembles, but
we leave assessment of their usability to the future.

\paragraph{Image segmentation.} 
A different approach aimed at extracting visual features from maps is image segmentation. 
However, we decided to develop and implement the patchwork method first for two reasons.
First, patches are meaningful semantic units on historical maps. Because they are highly flexible in nature (their size can vary), they accommodate maps that may have been created with varying levels of accuracy (in relation to any feature's true position on Earth).
Second, it is easy to create large training and evaluation datasets of patches, one of the pre-requisites for high-quality predictive CV models. 
The resulting trained models can be applied to large-scale datasets within reasonable computation time. 
For image segmentation, the hand-labelling of each pixel is tedious and expensive compared to patch-level labelling.
Nevertheless, pixel-level segmentation has advantages:
the results are not bounded to a specific patch size, and some patch-level outputs could be reconstructed from the pixel-level information.

For the results presented in this paper, we used $\approx$62K annotated patches from 30 map sheets.
It is not clear how many map sheets would need to be annotated for an image segmentation method to perform similarly to our current models.
We leave the implementation of image segmentation methods in \textit{MapReader} and assessing the usability of these models for the future.

\section{Conclusions}
We presented \textit{MapReader}, a free, open-source software library written in Python.
\textit{MapReader} allows users with little or no CV expertise to work with large collections of maps.
We presented a case study focusing on British rail infrastructure and buildings as depicted in a collection of $\approx$16K nineteenth-century British OS maps ($\approx$30.5M patches), demonstrating how patches on their own and in combination with other patches or external datasets can offer new insights into deciphering the spatial patterns of industrial development in modern Britain. 
Using an image classification task at patch level transforms a common, indeed unsophisticated, CV method into a radically new way for historians and others to interact with maps.
Interrogating the national landscape in terms of attributes like railspace or building density sets the stage for CV-driven research based on extensive new datasets allowing us to identify patterns, and, with the comparison of maps over time, changes to the built and natural environments.
Finally, we open-source $\approx$62K expertly annotated patches with the hope that this dataset will foster further collaboration between the fields of CV, Machine Learning and History, as well as with libraries and archives curating series map collections.

\paragraph{Acknowledgements:} 
The map images and metadata used in this research are provided by the National Library of Scotland. 
The data is available for non-commercial re-use under CC BY-NC-SA 4.0 licence (see \url{https://maps.nls.uk/copyright.html#exceptions-os}).

We thank Mariona Coll Ardanuy, Jon Lawrence and Joshua Rhodes for discussions and feedback. 
We thank many members of the Living with Machines team for testing early annotations.
This work was supported by Living with Machines (AHRC grant AH/S01179X/1) and The Alan Turing Institute (EPSRC grant EP/N510129/1). 
Living with Machines, funded by the UK Research and Innovation (UKRI) Strategic Priority Fund, is a multidisciplinary collaboration delivered by the Arts and Humanities Research Council (AHRC), with The Alan Turing Institute, the British Library and the Universities of Cambridge, East Anglia, Exeter, and Queen Mary University of London.

{\small
\bibliographystyle{ieee_fullname}
\bibliography{egbib}
}

\clearpage

\appendix
\renewcommand\thefigure{\thesection.\arabic{figure}}
\renewcommand\thetable{\thesection.\arabic{table}}
\setcounter{figure}{0}
\setcounter{table}{0}

\section{Interactive annotation tool}
\label{app:annot_tool}

\begin{figure}[h!]
    \includegraphics[width=1.0\linewidth]{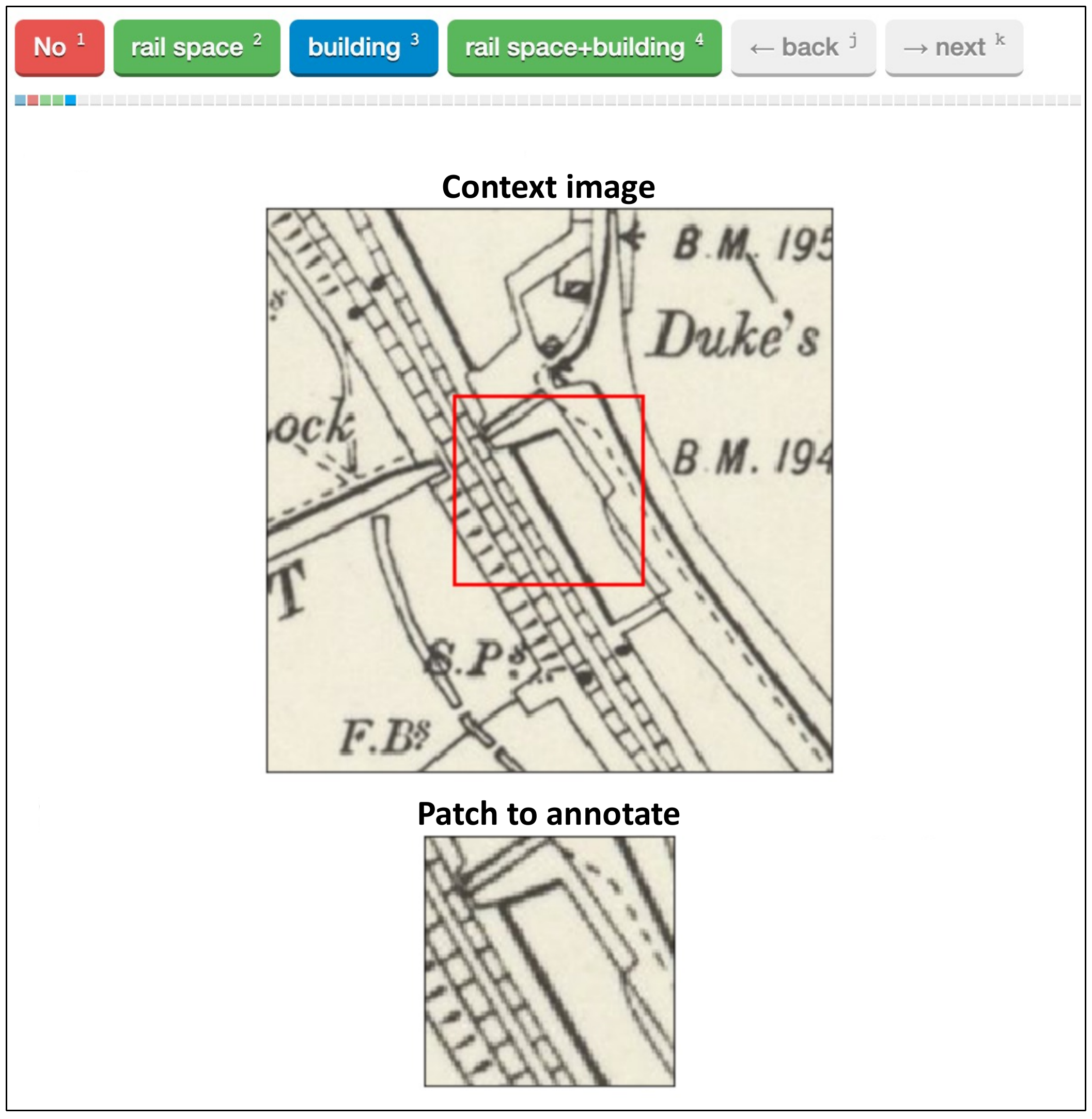}
    \caption{
    Interactive annotation tool of \textit{MapReader}.
    We developed this interface based on feedback from researchers from different disciplinary backgrounds, including historians, linguists, data scientists and software engineers.
    }
\label{fig:mr_annotation_interface}
\end{figure}

\textit{MapReader}'s interactive annotation tool provides a flexible way of annotating patches. The tool's flexibility lies in the simple options for controlling the interface.
For the patches on a given map sheet, the user can choose to see the context around a patch during annotation as shown in Fig.~\ref{fig:mr_annotation_interface}, and size of the context image can be set by the user.
The user can specify the labels (e.g., ``No'', ``rail space'', ``building'' and ``rail space + building'' in Fig.~\ref{fig:mr_annotation_interface}) in the configuration file or set them directly in the annotation interface which runs in a Jupyter Notebook. It is therefore very easy to experiment with using different labels on certain patch sizes, with or without the context image.
By default, \textit{MapReader} samples the patches randomly from a dataset.
However, the user can change the selection criteria, for example, the patches can be sampled according to their mean (or standard deviation) pixel intensities. This is helpful, for example, in identifying clusters of patches that are similar in terms of the distribution of printed visual features in a patch. Overall, the annotation tool is a place for users to work closely with the map images presented as patches. At the early stages of research design, this allows users to consider how labels relate to patches, and to revise patch size, label terms, or context image appearance as desired.

\section{MapReader interface}
\label{app:mr_python_interface}

\textit{MapReader} is an end-to-end CV pipeline, and it is available as a Python library. 
As an example, loading maps, slicing them into patches of size 50$\times$50 pixels and plotting four sample results can be executed by:

\begin{python}
from mapreader import loader

mymaps = loader(path_images)

# "method" can be pixel or meters
mymaps.sliceAll(path_save=path_save_patches, 
                slice_size=50,
                method="pixel")
                
mymaps.show_sample(4)
\end{python}

\noindent and initializing a CV classifier object using a pretrained ``resnet152'' model and showing the details of each layer (i.e., the number of trainable weights and biases, the name of each layer as specified in a model and the total number of trainable or ``frozen'' parameters) can be initiated with:

\begin{python}
from mapreader import classifier

myclassifier = classifier()
myclassifier.initialize_model("resnet152", 
                              pretrained=True)
                              
myclassifier.model_summary(only_trainable=False)
\end{python}

\noindent Other functionalities, such as retrieving maps from webservers, reading annotations, training, fine-tuning and model evaluation have similar easy-to-use interfaces. Consult \textit{MapReader}'s GitHub page for additional information and examples.

\section{Model performance on validation set}
\label{app:compare_valid_set}

\begin{table*}
\caption{
Performance of CV classifiers on the validation set with 12,404 labelled patches (20\% of all manually annotated patches). 
We used pretrained models from \textit{PyTorch Image Models} \cite{rw2019timm} and \textit{PyTorch} \cite{paszke2019pytorch}.
The models are ordered according to F1-macro. 
The F1-score for each label is also listed. 
F1-0: no [building or railspace]; F1-1: railspace; F1-2: building; and F1-3: railspace and [non railspace] building.
Time is for both training and validation.} \vspace{-0.5cm}
\begin{center}
\begin{tabular}{lrrrrrrrr}
\toprule
                     Model &  F1-macro &  F1-micro &  F1-0 &  F1-1 &  F1-2 &  F1-3 &  Time (m) / epoch &  \#params (M) \\
\midrule
            resnest101e \cite{zhang2020resnest} &     \textbf{96.74} &     \textbf{99.59} & \textbf{99.88} & 95.59 & \textbf{97.75} & 93.75 &              98.2 &         46.2 \\
  tf\_efficientnet\_b3\_ns \cite{tan2020efficientnet} &     96.50 &     99.52 & 99.84 & \textbf{95.67} & 97.12 & 93.37 &              89.7 &         10.7 \\
     resnest50d\_4s2x40d &     96.47 &     99.42 & 99.77 & 95.17 & 96.48 & \textbf{94.46} &              86.6 &         28.4 \\
  swsl\_resnext101\_32x8d \cite{DBLP:journals/corr/abs-1905-00546} &     96.33 &     99.42 & 99.79 & \textbf{95.67} & 96.39 & 93.47 &             130.8 &         86.8 \\
              resnet152 \cite{he2015deep} &     95.80 &     99.29 & 99.73 & \textbf{95.67} & 95.30 & 92.51 &              94.9 &         58.2 \\
resnest101e\_no\_pretrain &     90.67 &     98.23 & 99.25 & 85.97 & 89.47 & 88.00 &              98.3 &         46.2 \\
      swin\_base\_patch4 \cite{liu2021swin} &     77.05 &     93.39 & 96.59 & 64.54 & 68.02 & 79.04 &             109.9 &         86.7 \\
  vit\_base\_patch16\_224 \cite{dosovitskiy2020image} &     64.78 &     91.43 & 95.93 & 43.38 & 64.61 & 55.21 &              99.6 &         85.8 \\
\bottomrule
\end{tabular}
\end{center}
\label{table:models_valid_set}
\end{table*}

We used 37,212 patches for training (60\% of the total number of labelled patches), 12,404 (20\%) for validation and 12,404 (20\%) for testing the model performance.
For each neural network architecture, the model with the least validation loss was selected. Table~\ref{table:models_valid_set} lists the performance of these models as measured by F1 scores (macro, micro and for each label separately).
Based on these results, we selected ``resnest101e'' model for the results of this paper. 
See Table~\ref{table:test_models_01} for the performance scores of these models on the test set.

\section{Linking and enriching railspace data}
\label{app:linking_stops}

\begin{figure}
    \centering
    \includegraphics[width=1.0\linewidth]{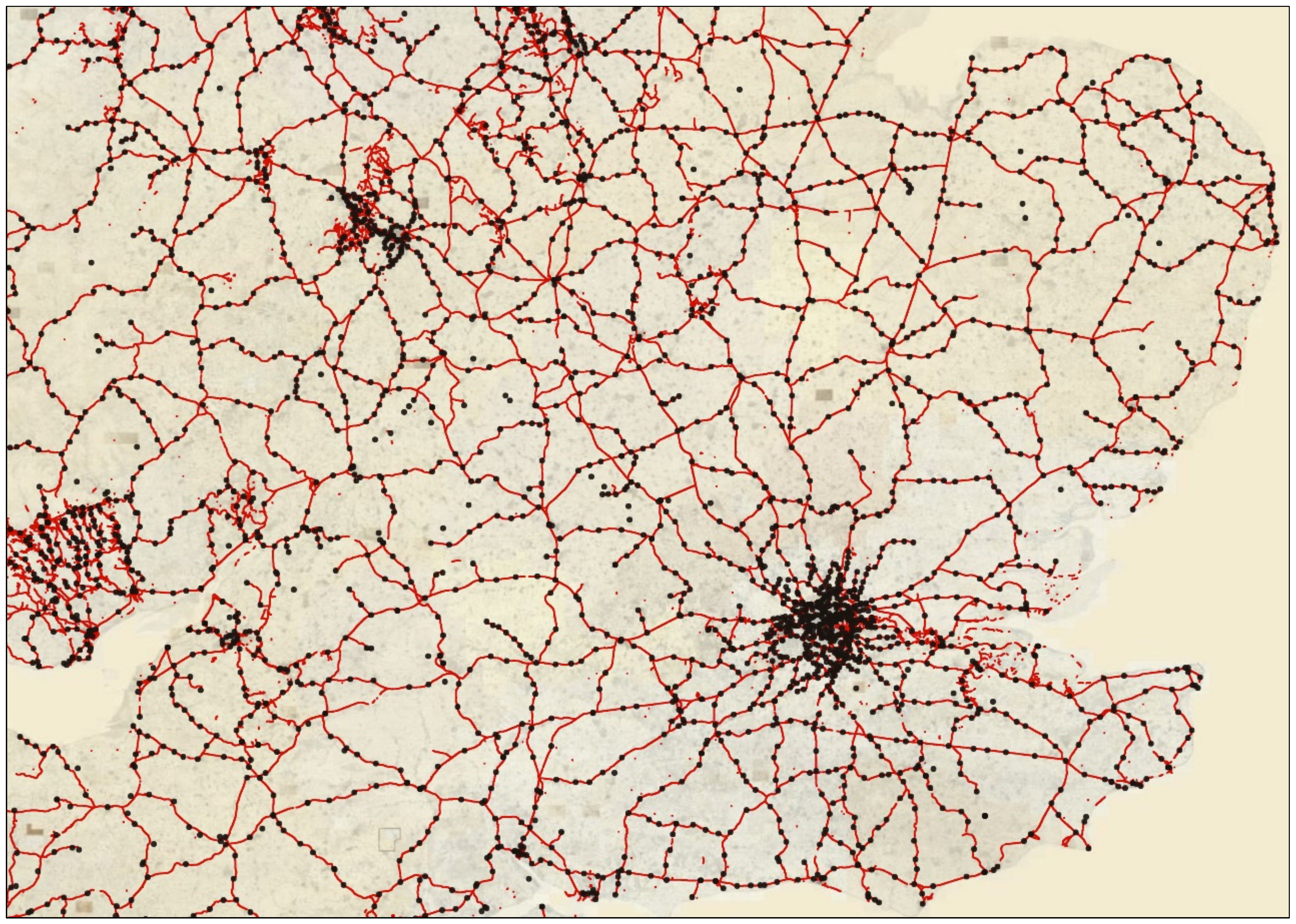}
    \caption{Visual comparison between railspace as predicted by \textit{MapReader} (red) and the location of railway stations from \textit{StopsGB} (black) \cite{lwm-station-to-station-2021}.} 
\label{fig:mr_cv_quicks}
\end{figure}

\textit{StopsGB} \cite{lwm-station-to-station-2021} is a structured dataset of over 12,000 railway stations based on Michael Quick's reference work \textit{``Railway Passenger Stations in Great Britain: a Chronology''}. In \cite{lwm-station-to-station-2021}, the authors used traditional parsing techniques to convert the original document into a structured dataset of stations, with attributes containing information such as operating companies and opening and closing dates. Moreover, for each station, a set of potential Wikidata candidates was identified using DeezyMatch \cite{hosseini-etal-2020-deezymatch,10.1145/3397536.3422236}. The best matching entity was then determined by using a supervised classification approach. Because most stations are linked to the best Wikidata candidate, this is a high-quality source for location data for historical British railway stations, an integral element within the class of railspace.

Fig.~\ref{fig:mr_cv_quicks} shows a visual comparison between \textit{StopsGB} and railspace as predicted by \textit{MapReader}. 
89.3\% of \textit{StopsGB} stations were within 150 meters of the center of a railspace patch (here, each patch has size of $\approx$100m$\times$100m). 

\section{Classification of buildings by their neighboring patches}

\begin{figure}
    \centering
    \includegraphics[width=1.0\linewidth]{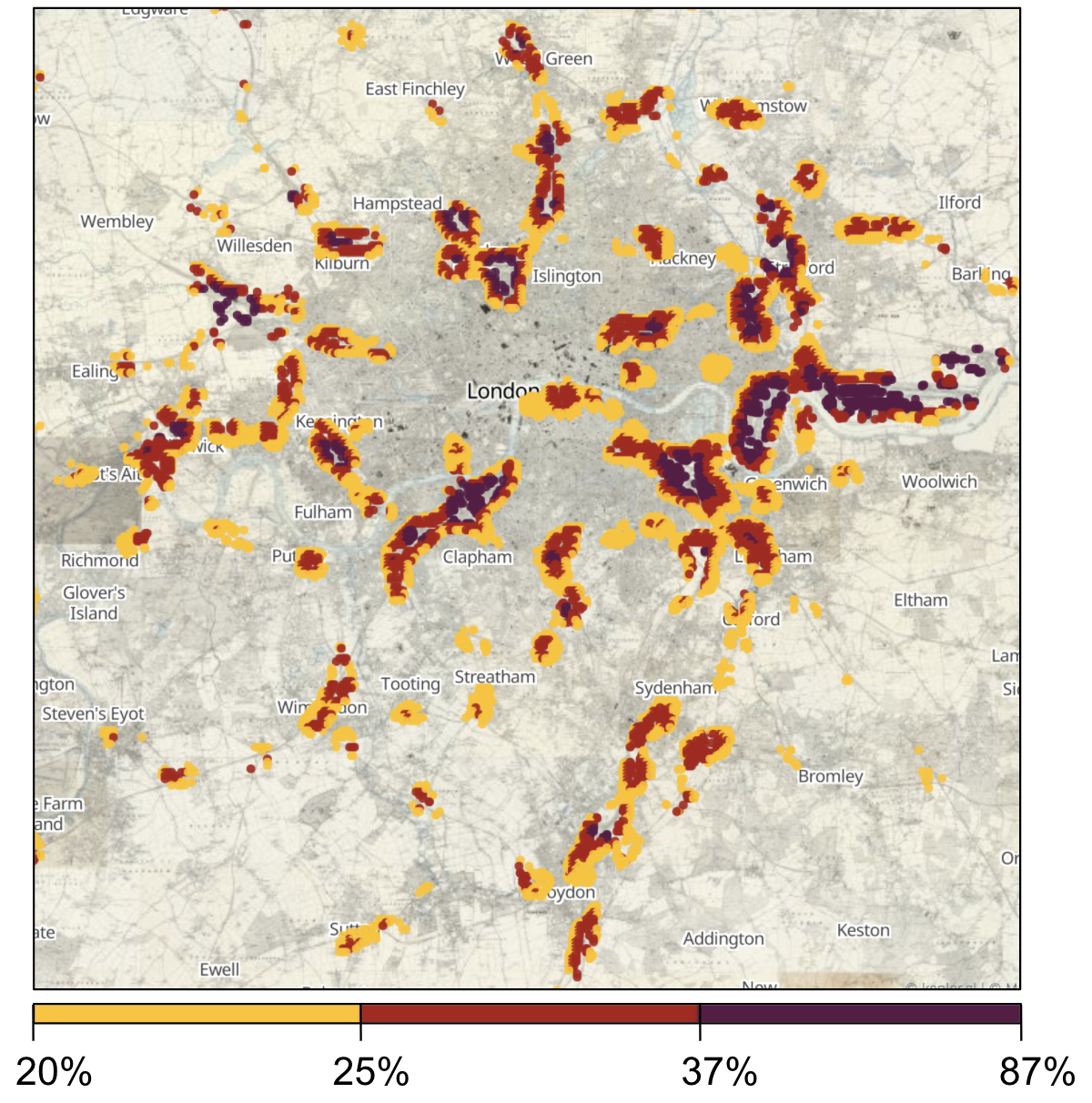}
        \caption{
        Building patches for which at least 20\% of the neighboring patches are classified as railspace.
        Neighbors are defined as all patches within 500 meters from a building patch. 
        Colour determined by quantiles (i.e., equal number of patches in each color interval). To make the legend legible, some have been merged in the visible colorbar.
        (The interval bounds, before being merged, were 20.0, 21.4, 23.2, 25.0, 27.5, 30.9, 37.0, 87.0\%.)
        }
\label{fig:london_railspace}
\end{figure}

\begin{figure}
    \centering
    \includegraphics[width=1.0\linewidth]{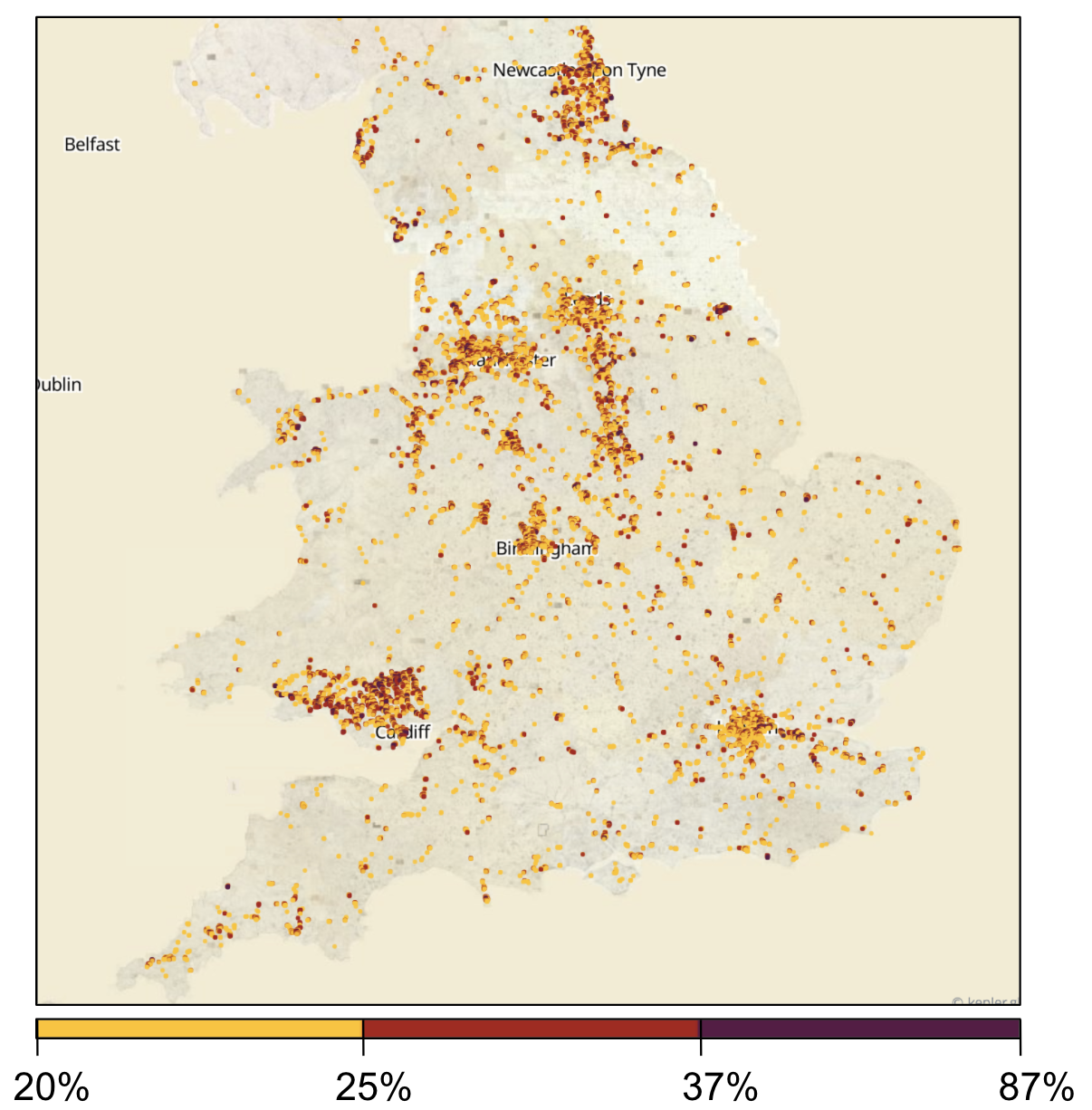}
    \caption{
    Similar to Fig.~\ref{fig:london_railspace} except that it shows the results for all of England and Wales.}
\label{fig:england_railspace}
\end{figure}

\begin{figure}
    \centering
    \includegraphics[width=1.0\linewidth]{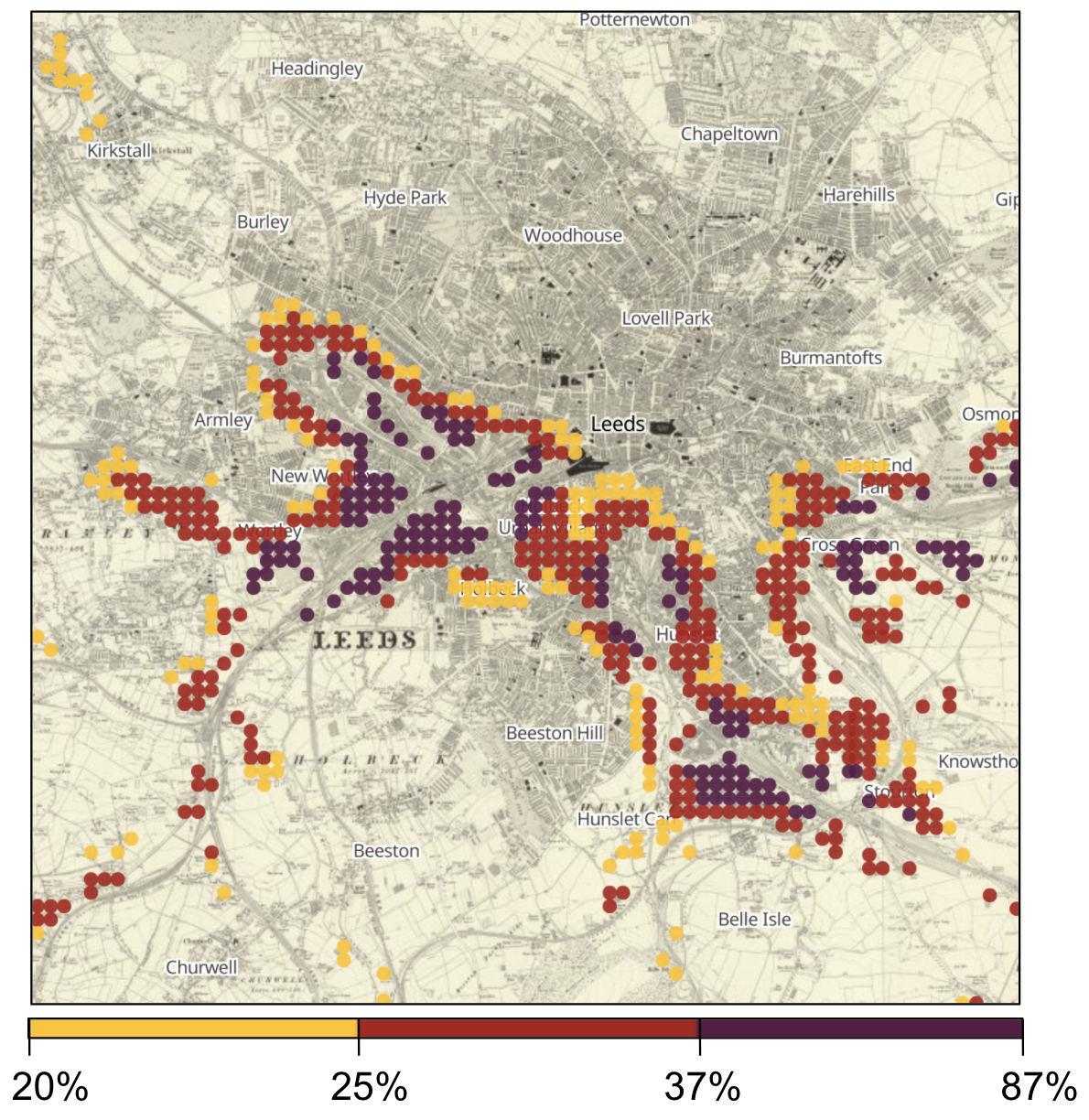}
    \caption{Similar to Fig.~\ref{fig:london_railspace} except that it shows results for greater Leeds.}
\label{fig:leeds_railspace}
\end{figure}

\textit{MapReader} produces data at patch level.
However, the context around each patch can be used to enhance the spatial semantics of our analysis. We do this by identifying space in terms of the \textit{relation} between (and not just the presence of) our labels (as well as other external datasets, like \textit{StopsGB}).
In Figs~\ref{fig:london_railspace}, \ref{fig:england_railspace} and \ref{fig:leeds_railspace}, we show how building and railspace patches relate to each other. 
Such a perspective (shown here for different regions and zoom levels) enables historians to search for places and compare them along new axes; for instance in this example, by identifying (first visually and then quantitatively) those areas that share certain features of London's railspace, expressed as the relation between given classes (such as ``building density'' and ``distance from a station''). 

In this way, \textit{MapReader} can help researchers go beyond the existing literature in their fields: on the one hand, by exploring the extent to which well-known case studies are in fact representative of wider datasets; and on the other hand, to help them identify entirely new areas of interest not previously considered, using visual semantic patterns as their guide.

\end{document}